\documentclass[letterpaper]{article} % DO NOT CHANGE THIS
\usepackage{aaai25}  % DO NOT CHANGE THIS
\usepackage{times}  % DO NOT CHANGE THIS
\usepackage{helvet}  % DO NOT CHANGE THIS
\usepackage{courier}  % DO NOT CHANGE THIS
\usepackage[hyphens]{url}  % DO NOT CHANGE THIS
\usepackage{graphicx} % DO NOT CHANGE THIS
\usepackage{natbib}  % DO NOT CHANGE THIS AND DO NOT ADD ANY OPTIONS TO IT
\usepackage{caption} % DO NOT CHANGE THIS AND DO NOT ADD ANY OPTIONS TO IT
\usepackage{algorithm}
\usepackage{algorithmic}
\usepackage{comment}

\usepackage{graphicx}
\usepackage{bm} 
\usepackage{amsthm,amsmath,amssymb}

\newtheorem{theorem}{Theorem}

\usepackage{mathrsfs}
\usepackage{multirow}
\usepackage{booktabs}  
\makeatletter
\renewcommand\@makefnmark{\hbox{\textsuperscript{\textdagger}}}
\renewcommand\@makefntext[1]{\noindent\hbox{\textsuperscript{\textdagger}}#1}
\makeatother

\usepackage{newfloat}
\usepackage{listings}
\DeclareCaptionStyle{ruled}{labelfont=normalfont,labelsep=colon,strut=off} % DO NOT CHANGE THIS
\floatstyle{ruled}
\newfloat{listing}{tb}{lst}{}
\floatname{listing}{Listing}
\title{Symbolic Neural Ordinary Differential Equations}
\author{
	Xin Li \textsuperscript{\rm 1},
	Chengli Zhao \textsuperscript{\rm 1}\thanks{Corresponding author},
	Xue Zhang \textsuperscript{\rm 1}$^\dag$,
	Xiaojun Duan \textsuperscript{\rm 1}
}
\affiliations {
	\textsuperscript{\rm 1}College of Science, National University of Defense Technology, Changsha, Hunan 410073, China\\
	li\_xin@nudt.edu.cn, chenglizhao@nudt.edu.cn, xuezhang@nudt.edu.cn
}

\usepackage{bibentry}
\begin{document}

\maketitle

\begin{abstract}
Differential equations are widely used to describe complex dynamical systems with evolving parameters in nature and engineering. Effectively learning a family of maps from the parameter function to the system dynamics is of great significance. In this study, we propose a novel learning framework of symbolic continuous-depth neural networks, termed Symbolic Neural Ordinary Differential Equations (SNODEs), to effectively and accurately learn the underlying dynamics of complex systems. Specifically, our learning framework comprises three stages: initially, pre-training a predefined symbolic neural network via a gradient flow matching strategy; subsequently, fine-tuning this network using Neural ODEs; and finally, constructing a general neural network to capture residuals. In this process, we apply the SNODEs framework to partial differential equation systems through Fourier analysis, achieving resolution-invariant modeling. Moreover, this framework integrates the strengths of symbolism and connectionism, boasting a universal approximation theorem while significantly enhancing interpretability and extrapolation capabilities relative to state-of-the-art baseline methods. We demonstrate this through experiments on several representative complex systems. Therefore, our framework can be further applied to a wide range of scientific problems, such as system bifurcation and control, reconstruction and forecasting, as well as the discovery of new equations.
\end{abstract}


\section{Introduction}
Complex dynamical systems, typically expressed as ordinary differential equations (ODEs) or partial differential equations (PDEs), are integral to scientific research and engineering applications across a wide range of domains \cite{wang2016data, brunton2022data, devaney2021introduction}. However, these systems are often subject to environmental changes during their evolution, which can lead to temporal and spatial variations in internal parameters or external perturbations \cite{wang2021learning}. This yields significant challenges when exploring the underlying dynamic mechanisms and predicting the behavior of the system in new scenarios. To address this issue, we aim to construct a surrogate model from observed data generated by parametric dynamic systems and the parameter functions \cite{benner2015survey,li2020neural,li2020fourier}. This model should be capable of approximating the operator mapping from the parameter functions to the system states. 

In most cases, we can only obtain the experimentally measured data from complex systems without prior knowledge of the underlying dynamical equations. Thus, data-driven methods are becoming increasingly important, including the autoregressive model \cite{navarro2008arma}, recurrent neural networks \cite{van2020review, cho2014learning, zhu2023leveraging, de2019gru}, graph neural networks \cite{di2022graph, pilva2022learning, liu2022multivariate}, and neural ODEs (NODEs) \cite{chen2018neural, holt2022neural, finlay2020train, bilovs2021neural}. In addition, recent work has shown that introducing physical priors can significantly enhance the learning performance of neural networks \cite{raissi2019physics, sanchez2019hamiltonian, cranmer2020lagrangian}. For instance, the PDE-NET proposed by Long \textit{et al.} \cite{long2018pde, long2019pde} utilizes the special convolution kernels to represent potential spatial derivative terms and adopts the idea of sparse identification to learn the nonlinear dynamics (SINDy) \cite{wu2019learning, Rudy2016DatadrivenDO}. The message-passing PDE solvers (MP-PDE) proposed by Brandstetter \textit{et al.} \cite{Brandstetter2022MessagePN} leverages a combination of traditional numerical methods to predict PDE systems. However, these methods often model only a specific system and require retraining when environmental parameters change.

In recent years, neural operators have gained significant attention \cite{Kovachki2021NeuralOL, Lu2021ACA, Jin2022MIONetLM, Chen2023LaplaceNO, Xiong2023KoopmanNO}. Among them, the Deep Operator Network (DeepONet, DON) \cite{Lu2019DeepONetLN} and Fourier Neural Operator (FNO) \cite{Li2020FourierNO} are two of the most prominent works. These methods encode the input function at fixed sample points and use a dedicated neural network architecture to directly predict the system state. Compared to traditional methods, these methods solve parametric dynamics systems faster and improve prediction accuracy during testing due to their grid-independent nature. However, they lack interpretability, do not explain dynamics well, and cannot utilize physical priors. Furthermore, they require a large amount of training data and have limited extrapolation capabilities outside the distribution of the training set.

To overcome the aforementioned challenges, we propose a novel learning framework, referred to as Symbolic Neural Ordinary Differential Equations (SNODEs), which establish sysbolic continuous-depth neural networks (SCDNNs) to effectively and accurately learn a family of dynamical systems. Our framework has several key advantages:
\begin{itemize}
	\item SNODEs enable accurate resolution-invariant reconstruction and prediction of the system state at any given time as well as any spatial grid (for PDEs), yielding a highly flexible and adaptable framework.
	
	\item SNODEs adopt a gradient flow matching pre-training strategy, which effectively circumvents the challenges associated with high computational loss and susceptibility to local optima inherent in traditional training method for SCDNNs.
	
	\item Our framework exhibits strong interpretability and extrapolation capabilities. This can be attributed to the effective implementation of operator learning and residual capture through a three-stage training approach in SNODEs, which leverages the strengths of both symbolism and connectionism.
\end{itemize}

\section{Related Work}
Our proposed framework builds upon the Neural Ordinary Differential Equations (NODEs) \cite{chen2018neural}. This method has gained significant attention in recent years as a continuous approximation of Residual neural Networks (ResNets)~\cite{he2016deep}. In ResNets, the $n$-th residual block transforms the hidden layer state from $\bm{h}_n$ to $\bm{h}_{n+1}$ using a trainable parameter vector $\bm{\theta}_n$ and a dimension-preserved function $\bm{f}(\bm{h}_n, \bm{\theta}_n)$, given by the following equation: 
\begin{equation*}
	\bm{h}_{n+1} = \bm{h}_n + \Delta t \cdot \bm{f}(\bm{h}_n, \bm{\theta}_n),
\end{equation*}
where the step size $\Delta t=1$. This discrete dynamical system can be viewed as the Euler discretization of the following ODE $\dot{\bm{h}}(t) = \bm{f}(\bm{h}(t),\bm{\theta})$, when we reuse the parameter vectors $\bm{\theta}_n$ as the parameter vector shared $\bm{\theta}$, and the number of residual blocks approaches infinity, and the step size approaches zero. Given an initial value $\bm{h}(t_0)$, one can obtain the state at any time using an ODE solver:
\begin{equation*}
	\begin{split}
		\bm{h}(t_1)
		= \bm{h}(t_0) + \int_{t_0}^{t_1} \bm{f}(\bm{h}(t),\bm{\theta})\mathrm{d}t,
	\end{split}
\end{equation*}
which can be recorded as $\text{ODESolve}\left[\bm{h}(t_0), t_0, t_1\right]$. Although recent works have extended this framework to various types of differential equations, including neural delay differential equations \cite{zhu2021neural}, neural controlled differential equations \cite{Kidger2020NeuralCD}, neural integro-differential equations \cite{Zappala2022NeuralIE}, neural stochastic differential equations \cite{Liu2019NeuralSS}, augmented neural ODEs \cite{Dupont2019AugmentedNO}, and stiff neural ODEs \cite{Kim2021StiffNO}, there has been limited research on applying the framework to parametric differential equations, particularly parametric PDEs.

\section{Method}
Parametric dynamic systems frequently involve a time and/or space-varying parameter function $\bm{u}(\bm{x},t)$. To model this family of systems, we present a novel symbolic continuous-depth neural network framework,  called symbolic NODEs (SNODEs). This framework is capable of effectively learning an operator mapping from the parameter function $\bm{u}(\bm{x},t)$ to the system state $\bm{s}(\bm{x},t)$. 

\subsection{The Framework of SNODEs}
Consider a dynamical system of the following general form:
\begin{equation} \label{eq_general}	
	\begin{split}
		&\partial_t \bm{s} = \bm{F}[\bm{u}, \bm{x}, t, \bm{s}, \bm{s}_{(x_i)}, \bm{s}_{(x_ix_j)},\cdots], \\
		&\qquad\quad \bm{s}(\bm{x},0) = \bm{s}_0, \quad \bm{x}\in \Omega,
	\end{split}
\end{equation}
where space domain $\Omega \subset \mathbb{R}^d $, system state $\bm{s}\in \mathbb{R}^{d_s}$, and $t\in [0,T]$. When $d=0$, the system is an ODE with respect to the parameter function $\bm{u}(t)$. When $d\geq 1$, it is a PDE system with respect to the parameter function $\bm{u}(\bm{x},t)$, which satisfies the periodic boundary conditions. Here, $\bm{s}_{(x_i)}$ represents all first-order partial derivatives of state $\bm{s}$ with respect to spatial variables, and $\bm{s}_{(x_ix_j)}$ represents all second-order partial derivatives, extending to higher-order derivatives up to the $q$-th order. In this paper, we focus on the scenario $q\leq 4$. We note that the proposed framework can be extended to higher-dimensional cases.

\begin{figure*}[t]
	\centering
	\includegraphics[width=0.90\textwidth]{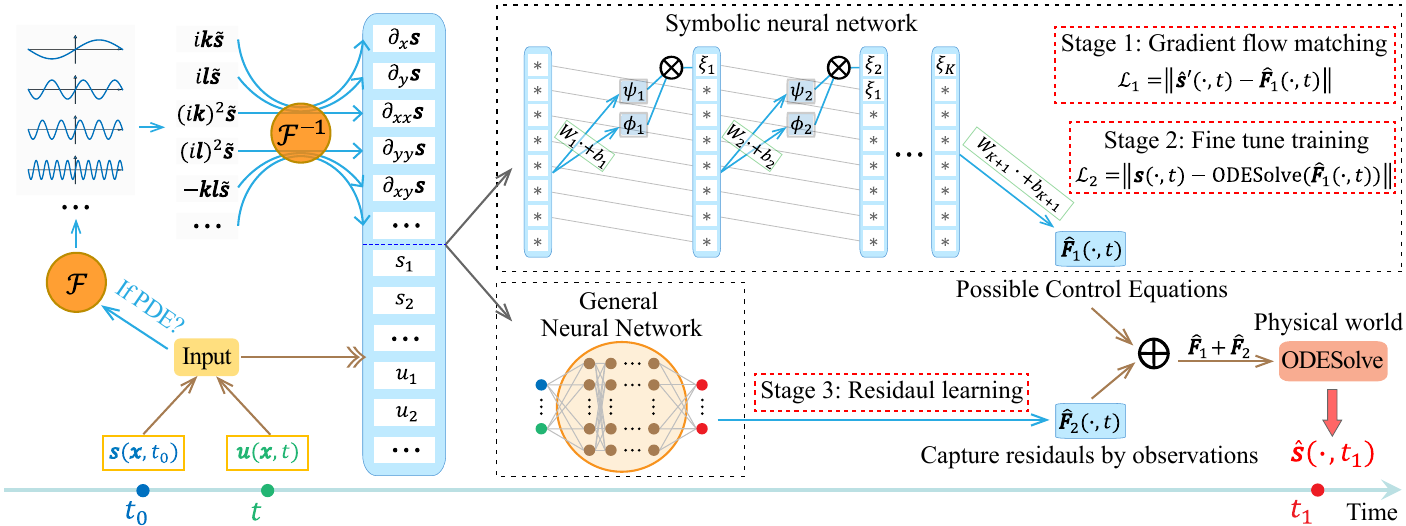} % Reduce the figure size so that it is slightly narrower than the column.
	\vspace{-0.2cm}
	\caption{The sketched framework of SNODEs. This framework, which includes the SymNet and GeNN components, takes state variables, parametric functions, and all possible spatial partial derivatives as inputs. It models a family of parametric dynamical systems using the proposed three-stage and adaptive learning strategy.}
	\label{model}
	\vspace{-0.35cm}
\end{figure*}

Our proposed framework enhances the accuracy of modeling the parametric dynamical systems by leveraging the information in both temporal and spatial domains.  For the temporal domain, as illustrated in Figure \ref{model}, SNODEs incorporate the parameter function $\bm{u}(\bm{x},t)$ as an additional input into a symbolic neural network (SymNet) and a general neural network (GeNN), and then numerically obtain the system state value at any given time $t_1$ by an ODE solver,
\begin{equation}\label{E_ODEsolve}
	\begin{split}
		\bm{s}(\bm{x},t_1) &= \text{ODESolve}[\bm{s}(\bm{x},t_0), t_0, t_1,\bm{u}(\bm{x},t)]\\
		&=\bm{s}(\bm{x},t_0)+\int_{t_0}^{t_1}\bm{\hat{F}}[\bm{s}(\bm{x},t),\bm{u}(\bm{x},t),t]\mathrm{d}t,
	\end{split}
\end{equation}
where $\bm{\hat{F}} = \bm{\hat{F}}_1 + \bm{\hat{F}}_2$ represents the vector field estimated by the sum of two neural networks. During the ODEsolve process, one can select an appropriate numerical integration to explicitly make a trade-off between the numerical precision and the computational cost.

%such as the fully connected neural networks, convolutional neural networks, and graph neural networks. 

To address the spatial dimensions in PDE systems, we consider a 2-dimensional PDE with periodic boundary conditions. In this case, directly learning the spatial partial derivatives of the system state from the neural network in Eq.~\eqref{E_ODEsolve} is often challenging and requires the incorporation of additional physical priors. To overcome this challenge, we propose a novel strategy for the integration of the spatial derivative terms. Specifically, our approach involves first applying a two-dimensional Fourier transform $\mathscr{F}$,
\begin{equation}\label{E_fourier}
	\bm{\tilde{s}}(\tilde{x},\tilde{y},t)=\mathscr{F}[\bm{s}(x,y,t)] = \int\int_{\Omega} \bm{s}(x,y,t)e^{-2\pi i(x\tilde{x}+y\tilde{y})},
\end{equation}
where $(x,y)\in \Omega$ and $t\in[0,T]$, and $\mathscr{F}^{-1}$ is the inverse transform of $\mathscr{F}$. According to the differential property of the Fourier transform, the partial derivative of the system state with respect to the space can be calculated as follows:
\begin{equation*}
	\begin{split}
		&\mathscr{F}\left(\frac{\partial^{n+m} \bm{s}}{\partial^n x\partial^my}\right)(\tilde{x},\tilde{y})=(2\pi i\tilde{x})^n(2\pi i\tilde{y})^m \mathscr{F}(\bm{s}), \\
		&\frac{\partial^{n+m} \bm{s}}{\partial^n x\partial^my} = \mathscr{F}^{-1} \left[(2\pi i\tilde{x})^n(2\pi i\tilde{y})^m \mathscr{F}(\bm{s}) \right] .
	\end{split}
\end{equation*} 
Given that the right-hand side $\bm{F}$ of most PDEs consists of only a few interacting symbols (see Eq.~\eqref{E_ODEsolve}), incorporating additional physical prior information is possible. 

To achieve this, we utilize the concept of the symbolic regression to effectively learn the underlying dynamics. Drawing inspiration from the Symbolic Neural Network (SymNet) \cite{sahoo2018learning, long2019pde}, we propose a continuous-depth multi-layer SymNet. In the first layer, we present the first-order term of all possible symbols, denoted as $\bm{L}_1=[\bm{u},\bm{s}, \partial_x \bm{s}, \partial_y \bm{s}, \partial_{xx} \bm{s}\cdots]^\top$. The length of $\bm{L}_1$ depends on the spatial dimension and the highest order of partial derivatives. In the next layer, we define 
\begin{equation}\label{E_symbol}
	\bm{L}_{k+1} = [\bm{\psi}_k\times \bm{\phi}_k, \bm{L}_k^\top]^\top, \, [\bm{\psi}_k,\bm{\phi}_k]^\top=\bm{W}_k \bm{L}_k+ \bm{b}_k,
\end{equation}
where $k\in \{1,2,\cdots, K\}$, $K$ is the number of hidden layers within SymNet, $\bm{W}_k$ and $\bm{b}_k$ are the trainable parameters. It allows us to use $\bm{\psi}_1\times \bm{\phi}_1$ to learn all quadratic product terms of elements from the layer $\bm{L}_1$. Our multi-layer SymNet can be extended to learn arbitrary order polynomials of elements from $\bm{L}_1$, thereby enabling the learning of the dynamics $\bm{F}$. The SymNet in Figure \ref{model} illustrates the execution process of a $K$-layer SymNet. Here, we need to consider the following discrete form via a proper discretization of the spatial domain,
\begin{equation*}
	\begin{split}
		&\bm{L_1}=[\bm{u},\bm{s},\mathscr{F}^{-1}(i\bm{k}\bm{\tilde{s}}),\mathscr{F}^{-1}(i\bm{l}\bm{\tilde{s}}),\mathscr{F}^{-1}((i\bm{k})^2\bm{\tilde{s}}),\cdots]^\top, \\ 
		&\bm{k}=\frac{2\pi}{N}(\bm{e}_N,\cdots,\bm{e}_N),\quad \bm{l} = \frac{2\pi}{M}(\bm{e}_M,\cdots,\bm{e}_M)^\top,
	\end{split}
\end{equation*} 
where $\bm{\tilde{s}} = \mathscr{F}(\bm{s})$, $\bm{e}_X$ denotes the vector $(0,1,\cdots,X-1)^{\top}$, $N$ and $M$ represent the number of sample points along the $x$ and $y$ dimensions, respectively. Additionally, the product between $\bm{k}$, $\bm{l}$, and $\bm{\tilde{s}}$ is obtained through the element-wise multiplication, and $\mathscr{F}$ and $\mathscr{F}^{-1}$ denote the discrete Fourier transform and inverse transform, respectively. 

Furthermore, we also introduce a GeNN for residual learning. This GeNN can embody various prevalent neural network architectures, such as fully connected neural networks, convolutional neural networks. Finally, using Eq.~\eqref{E_ODEsolve} and $\bm{\hat{F}}$, we can compute the value of $\bm{s}(x,y,t_1)$ at any time point $t_1$.

\subsection{Learning Strategies for SNODEs}
Utilizing Eq.~\eqref{E_ODEsolve} directly for training SNODEs encounters several challenges, including the high computational costs and the risk of converging to local optima associated with the classic NODE method, as well as the issue of gradient explosion. Therefore, we propose a three-stage training approach to efficiently model unknown dynamics, as shown in Figure \ref{model}.

In stage 1, we employ a strategy of gradient flow matching \cite{li2024fourier} to pre-train the SymNet. Here, we estimate the temporal gradient $\bm{\hat{s}}'$ through Fourier analysis
\begin{equation*}
	\begin{split}
		\bm{\hat{s}}' = \mathscr{F}_t^{-1} (i \bm{u} \mathscr{F}_t (\bm{s})), \, \bm{u} = \frac{2\pi}{N_t}(\bm{e}_{N_t}, \cdots, \bm{e}_{N_t}),
	\end{split}
\end{equation*}
where $N_t$ denotes the number of sampling points in the temporal direction, $\mathscr{F}_t$ is the Fourier transform along the temporal axis. Then the flow matching loss function $\mathcal{L}_1 = \|\bm{\hat{s}}' - \bm{\hat{F}}_1 \|$. In stage 2, we fine-tune the SymNet $\bm{\hat{F}}_1$ by minimizing the prediction error of ODESolve, and the loss function $\mathcal{L}_2 = \|\bm{s}-\text{ODESolve}(\bm{\hat{F}}_1)\|$. In stage 3, we keep SymNet fixed and further train GeNN for residual learning with the loss function $\mathcal{L}_3 = \|\bm{s}-\text{ODESolve}(\bm{\hat{F}}_1 + \bm{\hat{F}}_2) \|$. Moreover, to enhance the sparsity of the inferred SymNet network, L1 regularization was incorporated into both $\mathcal{L}_1$ and $\mathcal{L}_2$, with the regularization coefficient, $\alpha$, serving as a hyperparameter.

During the training process in stages 2 and 3, we employ an adaptive training strategy. To begin, we simply set the prediction steps to $1$ and increase it once the training error falls below a predefined threshold. Secondly, we incorporate an adaptive learning rate, which involves adjusting the learning rate at evenly spaced intervals across batches. Finally, we consider numerical methods for ODE solvers. In the early stages of training, simpler methods such as Euler should be used to expedite training (high-precision methods may cause numerical instability and training failure). In the later stages of training, higher-precision methods such as Runge-Kutta, adaptive-step solvers, can be employed to further train the model and achieve longer-term predictions on the test set. 

%The aforementioned training strategies can efficiently model a family of dynamic systems, aligning more closely with realistic physical world. 

%To begin, we introduce the concept of adaptive prediction steps, denoted as $t_p$. During the training process, we simply set $t_p$ to $1$ and increase it once the training error falls below a predefined threshold. Secondly, we incorporate an adaptive learning rate, which involves adjusting the learning rate at evenly spaced intervals across batches. Finally, we consider numerical methods for ODE solvers. In the early stages of training, simpler methods such as Euler should be used to expedite training, while high-precision methods may cause numerical instability and training failure during this period due to the large randomness. In the later stages of training, higher-precision methods such as Runge-Kutta, adaptive-step solvers, can be employed to further train the model and achieve longer-term predictions on the test set. 

\subsection{Theoretical Results}
The proposed framework effectively models the mapping operator from parametric functions to system states. To theoretically validate our method, here we present the Theorem \ref{universal} about universal approximation theorem for SNODEs (see Appendix A.1 for specific proof). Additionally, see Appendix C.1 for the extension of the SymNet.

\begin{theorem} \label{universal}
	(Universal approximation theorem for SNODEs.) 
	Suppose $\bm{F}$ (denoted by $\bm{F}(\bm{u},\bm{s})$ in Equation~\eqref{eq_general}) is any nonlinear function acting on $\bm{u}$ and at most $q$-th order space derivatives of $\bm{s}$, the functions $\bm{u}(\bm{x},t)$ and $\bm{s}(\bm{x},t)$ are defined on the domain $\Omega\subset\mathbb{R}^d$ with periodic boundary conditions with $0\le t\le T$. Then for any $\varepsilon>0$, there are positive integers $K$, weights $\bm{W}_k,~\bm{b}_k,~k=1,\cdots K$ such that 
	\begin{equation*}
		\begin{aligned}
			&\bm{L_{k+1}} = [\bm{\psi}_k\times \bm{\phi}_k, \bm{L}_{k}^\top]^\top,\, [\bm{\psi}_k,\bm{\phi}_k]^\top=\bm{W}_k \bm{L}_k+ \bm{b}_k, \\ &\bm{L_1}=[\bm{u},\bm{s},\partial_{x_1}\bm{s},\cdots,\partial_{x_d}\bm{s},\partial_{x_1x_1}\bm{s},\cdots]^\top,\\
			&\vert\bm{F}(\bm{u},\bm{s})-(\bm{W}_n\bm{L}_n+\bm{b}_n)\vert<\varepsilon,
		\end{aligned}
	\end{equation*}
	holds for all $\bm{x}=(x_1,\cdots,x_d)\in\Omega$,~$\bm{u}\in C(\Omega\times [0,T])$ and the corresponding solution $\bm{s}$ of the Equation~\eqref{eq_general} with the initial $\bm{s}(\cdot,0)\in C(\Omega)$. Here, $\bm{L}_1$ is the dictionary including all possible $q$-th order terms, $C(\Omega)$ is the Banach space of all continuous functions defined on $\Omega$ with norm $\Vert f\Vert_{C(\Omega)}=\max_{\bm{x} \in \Omega}\vert f(\bm{x})\vert$. 
\end{theorem}

In fact, the number of elements in the $\bm{L}_1$ layer of SymNet is a crucial parameter, and we can calculate it using Theorem \ref{cal_S} (see Appendix A.2 for specific proof). In most practical scenarios, it is typically observed that $d\leq 3$ and $q\leq 4$. Based on this, one can infer that the dimension of $\bm{L}_1$ is generally less than $35d_s+d_u$, and our improved training strategies demonstrate robustness under this condition.

\begin{theorem} \label{cal_S}
	Assuming the system state variable $\bm{s}(\bm{x},t)$ in Eq.~\eqref{eq_general} has a dimensionality of $d_s$ and a spatial dimensionality of $d$, and $\bm{F}$ represents at most $q$-th order space derivatives of $\bm{s}$. Then we can derive a recursive formula for the length $S_d(d_s,q)$ of $\bm{L}_1$ with respect to $d$, and it holds that
	\begin{equation*}
		S_d(d_s,q) =  
		\begin{cases}
			d_s (q + 1) +d_u, \,\, & d=1, \\
			\sum_{i=0}^q (S_{d-1}(d_s,i)-d_u) + d_u, \,\, & d>1.
		\end{cases}
	\end{equation*}
\end{theorem}

In addition, in the NODEs \cite{chen2018neural,zhuang2020adaptive}, the adjoint sensitivity method is used to calculate the gradients with a low cost of memory independent of the depth/time. Similarly, we can obtain the adjoint dynamics of the SNODEs, see Appendix A.3 for details.

\section{Experiments}
In this section, we provide a comprehensive analysis of the performance of our method across experiments with 64GB RAM and NVIDIA Tesla V100 GPU 16GB. To train and test our approach for a given dynamical system $\bm{s}(\bm{x},t,\bm{u}(\bm{x},t))$, we generate $N_\mathrm{tr}$ and $N_\mathrm{te}$ trajectories, respectively. Specifically, we utilize a Gaussian random field (GRF) to generate a $1$-d parameter function for each track. The GRF is characterized by its mean, denoted by $\mu$, and a radial basis function (RBF) kernel, given by
\begin{equation}
	u \sim \mathcal{G}\left\{\mu, \text{exp}\left[{||t_1-t_2||}^2/(2l^2)\right]\right\},
\end{equation}
where $l$ represents the length scale that determines the smoothness of the sampling function. We then multiply the sampling function by an output scaling factor $c$ to obtain our parameter function $u$. In practice, we can only sample a discrete set of points instead of a continuous function. In this case, we can obtain a continuous parameter function through interpolation methods, such as the cubic spline interpolation, as described in Appendix B.1. In addition, the parameter settings for all experiments are provided in Appendix B.2.

\subsection{A Simple 1-d Parametric ODE}
\begin{figure}[t]
	\centering
	\includegraphics[width=0.46\textwidth]{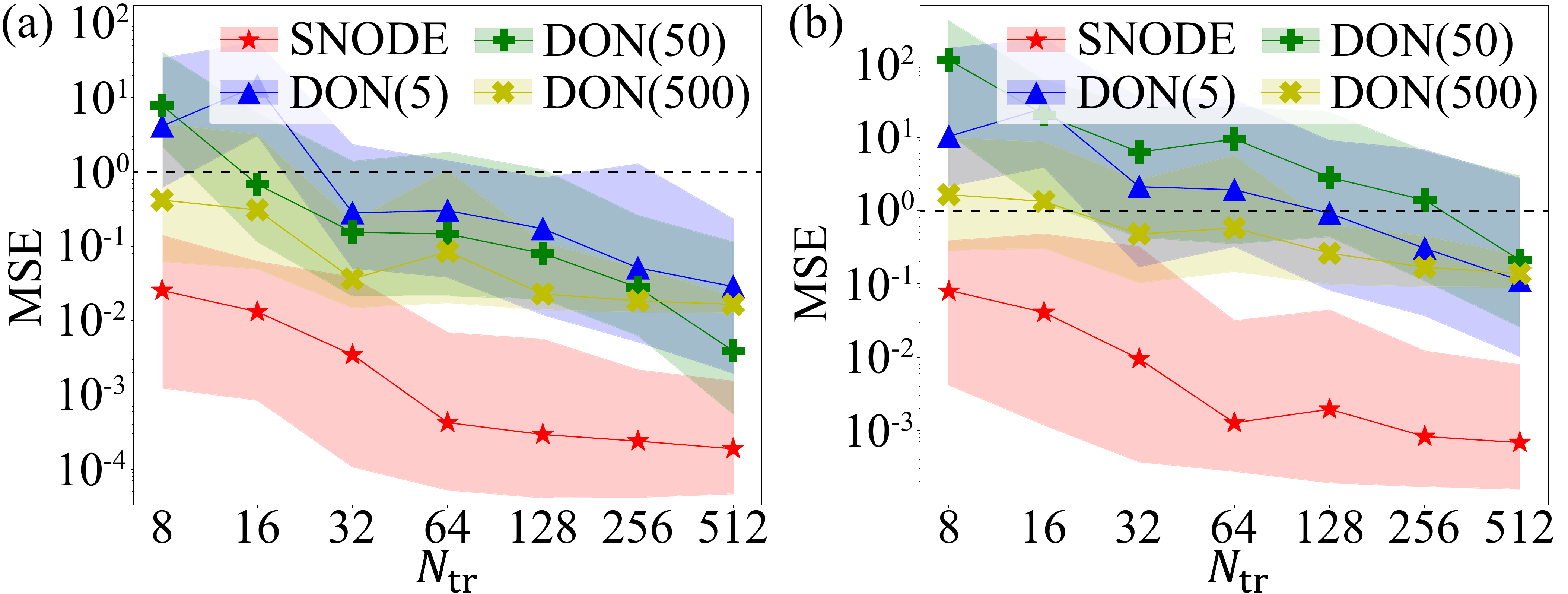}
	\vspace{-0.2cm}
	\caption{Operator learning for system \eqref{ode}. (a) The mean squared error (MSE) of different methods and different $N_\mathrm{tr}$ in the testing set. (b) The MSE in the extrapolation experiment. Here, $m$ in ``DON($m$)'' represents the number of the uniform sampling points of the parameter function.} 
	\label{F_1ode}
	\vspace{-0.35cm}
\end{figure}
We first consider a simple $1$-d ODE, described by
\begin{equation}\label{ode}
	\partial_t s(t) = F[s(t),u(t),t], \qquad t\in [0,1],
\end{equation}
where $u(t)$ represents a sampling function from a GRF on the interval $[0,1]$. We conduct experiments on the nonlinear example from the DeepONet work \cite{Lu2019DeepONetLN}, where $F[s(t),u(t),t]=-s^2(t)+u(t)$. To increase the task difficulty, we randomly sample the initial value $s_0$, the length scale $l$, and the output scale $c$ from their respective uniform distributions, as shown in the Appendix B.2.

\begin{figure*} 
	\centering
	\includegraphics[width=0.8\textwidth]{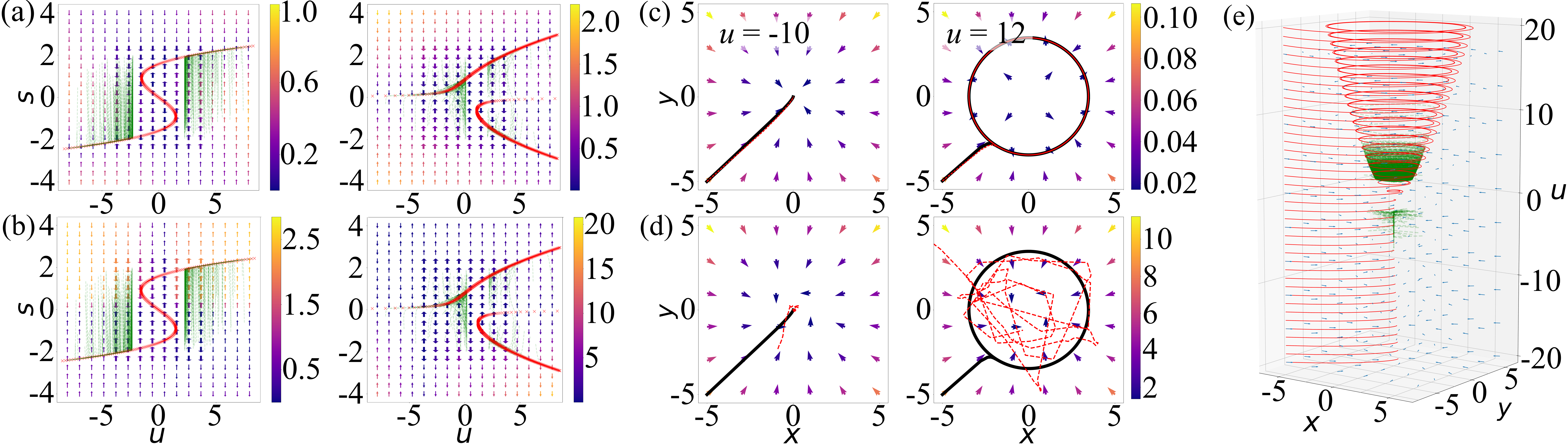}
	\vspace{-0.2cm}
	\caption{Experimental results in systems with saddle-node, pitchfork and Hopf bifurcations. (a), (c), (e) Predicting bifurcation dynamics using SNODEs. (b), (d) Predicting bifurcation dynamics using DeepONet.
	Here, the green dashed lines indicate the training data, the red dots represent the true bifurcation diagram, the arrows depict the predicted vector field, the black line represents the true trajectory, and the red line represent predicted trajectory.} 
	\label{F_snpf}
	\vspace{-0.35cm}
\end{figure*} 

The experimental results are presented in Figure \ref{F_1ode}. Our approach consistently outperforms the baselines regarding the prediction error, regardless of the number of points sampled from $u(t)$ by the DeepONet method. Moreover, our method exhibits a more significant advantage in the extrapolation prediction when the output scale $c$ selected exceeds the training setting ($c\in [0,10]$ for training and $c=14$ for the extrapolation experiment). This can be attributed to our approach's ability to fully utilize the temporal information of $u(t)$ within the framework of NODEs and learn the inherent relationship between the parameter function $u(t)$ and the system dynamics, thereby enabling highly accurate modeling of the parametric ODEs. 

%To further illustrate the potential application of our method in high-dimensional systems, we provide experimental results on the dynamics of a high-dimensional reservoir computing in Appendix C.1.

\subsection{Learning Dynamical Systems Near the Bifurcation}
In this section, we consider a ubiquitous and significant bifurcation scenario that has gained considerable attention across various scientific fields. When a parameter function reaches a bifurcation point, the qualitative or topological nature of the system undergoes a mutation, exhibiting highly complex dynamical behavior. By accurately learning the operator from the parameter function to the system state, SNODEs can facilitate the extrapolation prediction of the bifurcation dynamics of the system. This capability is vital for comprehending and predicting the system's tipping point.

We select three distinct dynamical systems with varying bifurcation types with respect to the parameter function $u(t)$, where $u(t)$ is sampled from the GRF near the bifurcation value $u^*$ of the system parameter. The first is the saddle-node bifurcation with $u^* = \pm 2.28$, and its dynamical equation is
$$\partial_t s(t) = u + \theta_1s + \theta_2s^3, \, \theta_1=2.5, \, \theta_2=-1. $$
The second is the pitchfork bifurcation with $u^* = 1.19$, which has the form
$$\partial_t s(t) = \theta_1 + us + \theta_2s^3, \, \theta_1=0.5, \, \theta_2=-1. $$
The finally is the Hopf bifurcation with $u^* = 0$, which reads
\begin{equation*} 
	\begin{split}
		& \partial_t s_1(t) = us_1 - s_2 - s_1(s_1^2+s_2^2), \\
		& \partial_t s_2(t) = s_1 + us_2 - s_2(s_1^2+s_2^2).
	\end{split}
\end{equation*}
Subsequently, we assess our SNODEs method by predicting these three bifurcation dynamics.

To begin with, we consider two $1$-d dynamical systems that exhibit a saddle-node bifurcation with a hysteresis loop, and a pitchfork bifurcation with a cusp catastrophe \cite{Szp2021ParameterIW}, respectively. To increase the task difficulty, we control the parameter function to avoid passing through a neighborhood near the bifurcation point, and then test the extrapolation ability of our method (the results are shown in Figure \ref{F_snpf}(a)). To verify the superiority of our method, we conduct the same experiments using the DeepONet as well. The results are presented in Figure \ref{F_snpf}(b). It is evident that the DeepONet method exhibits poor performance in the extrapolation prediction outside the green area. However, our method, with its powerful extrapolation prediction ability, accurately learns the bifurcation dynamics within a larger region.

Additionally, we consider a $2$-d dynamical system that exhibits Hopf bifurcation. To investigate the bifurcation dynamics of the system, we sample the parameter functions on both sides of the bifurcation value and generate the corresponding dynamical trajectories as the training set, as illustrated by the green dashed line in Figure \ref{F_snpf}(e). To discover the bifurcation dynamics of the system, we perform experiments on the parameter functions with a constant value in a wide range. As depicted in Figures \ref{F_snpf}(c)-(e), the red dashed line represents the predicted trajectory, the arrows depict the predicted vector field, and the black solid line corresponds to the true trajectory. The experimental results demonstrate that the learned operator mapping using our proposed framework exhibits superior capabilities in reconstructing and extrapolating the bifurcation dynamics.

The SNODEs framework, augmented with its efficient training strategies, has been successfully applied in the aforementioned parameteric ODE systems. When furnished with sufficient training data, the SymNet within the SNODEs framework accurately infers the underlying equations of unknown systems. In scenarios where the training data is scarce and the system is complex, SymNet can still precisely identify key components of the underlying dynamics. Following this identification, by integrating with GeNN's residual learning, it can realize a more accurate modeling of the system. For more training details, please refer to Appendix B.3.

\subsection{Applications in Parametric PDEs}
We further demonstrate the modeling capability of SNODEs in parametric PDEs with unknown equations, such as the diffusion-reaction (DR), Kuramoto-Sivashinsky (KS), and $2$-d Navier-Stokes (NS) systems. To leverage the physical prior knowledge, we incorporate the potential spatial derivatives as additional inputs to our SNODEs method, thereby enhancing the precision in learning the dynamics from parametric functions to system states. In the following experiments, we consider periodic boundary conditions, and the initial conditions and other experimental configurations are presented in Appendix B.2.

\begin{figure} 
	\centering
	\includegraphics[width=0.47\textwidth]{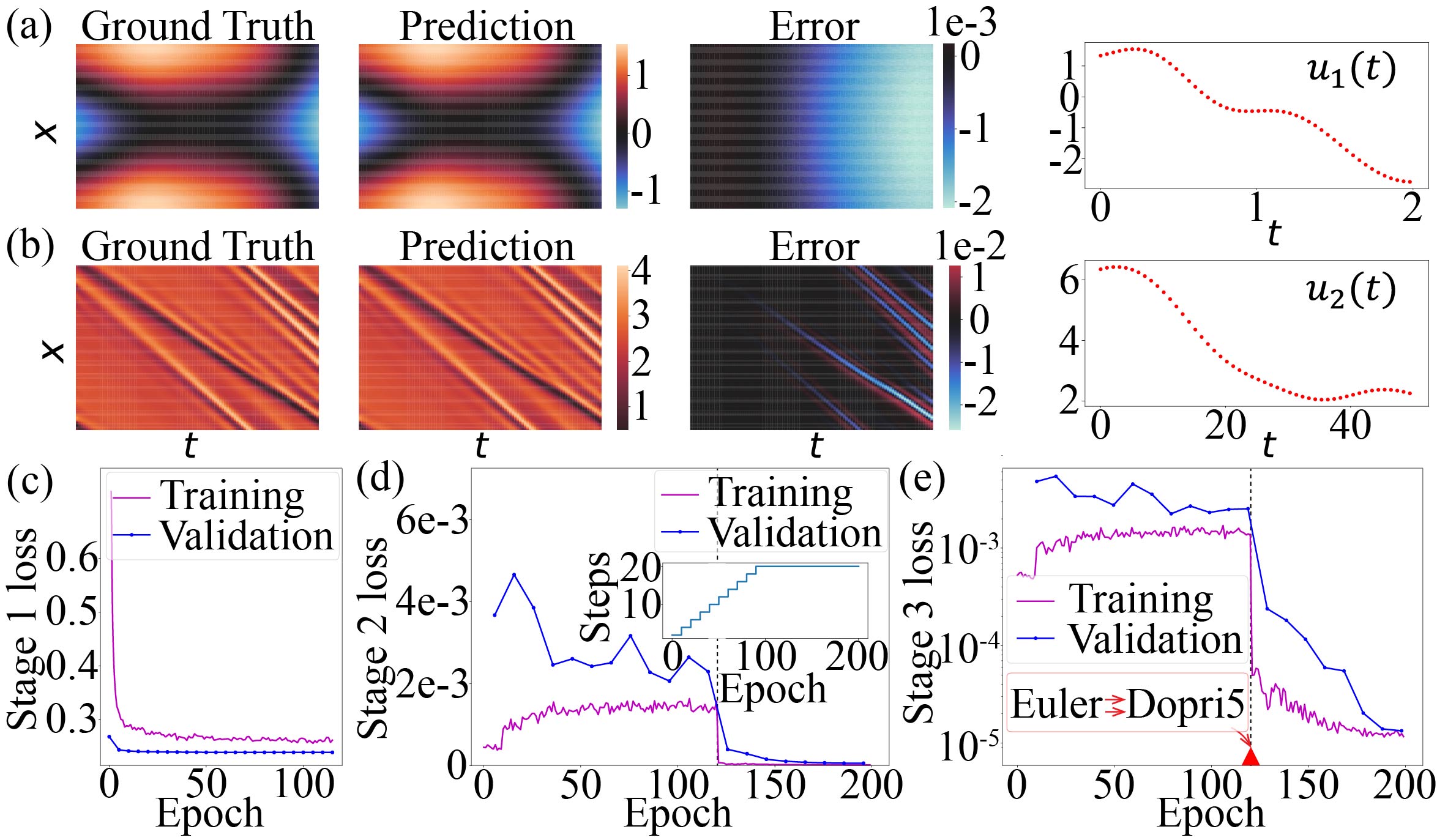}
	\vspace{-0.4cm}
	\caption{Predicting DR and KS systems using SNODEs. (a) and (b) show the testing examples for the DR and KS systems, respectively. Here, the training set features a spatial resolution of $N_x = 32$, whereas the test set has $N_x = 128$. (c), (d), and (e) respectively illustrate the variations in training and validation losses across three training stages.} 
	\label{F_dr_ks}
	\vspace{-0.4cm}
\end{figure}    

We first consider the DR system. In practice, the source term may vary over time due to changes in environmental factors, yielding the following equation with a time-varying source term $u(x,t)$: 
\begin{equation} \label{dr_equation}
	s_t = Ds_{xx}+Ks^2+u(x,t), \quad x\in[0,1],\, t\in[0,1], 
\end{equation}
where $D=0.01$ is the diffusion coefficient and $K=0.01$ is the reaction rate. Here, we use $u(x,t) = (\pi x)/5+u_1(t)$, where $u_1(t)$ is a sampling function from the GRF. In addition, we consider the KS equation, which exhibits complex chaotic dynamics, of the following form
\begin{equation} \label{ks_equation}
	\begin{split}
		s_t = -ss_x-s_{xx}-u(x,t)s_{xxxx}, \\
		x\in [0,32\pi],\, t\in[0,20],
	\end{split}
\end{equation}
where $u(x,t) = x/16+u_2(t)$, and $u_2(t)$ is the sampling function from the GRF. After training, our SNODEs framework exhibits excellent modeling capabilities on the aforementioned PDE systems. Moreover, our framework can be naturally applied to the super-resolution learning by using higher-resolution spatiotemporal data as the test set. In the temporal dimension, our approach is capable of estimating the system state for any given future moment through the ODESolve process. In terms of spatial dimension, our method allows for training with lower resolution data ($N_x=32$) and testing in higher resolution data ($N_x=128$). The corresponding outcomes are presented in Figures \ref{F_dr_ks}(a)-(b).

Here, we take the DR system as an example and provide the training details for our SNODEs framework. In stage 1, the rapid capture of critical components of unknown dynamics is facilitated through flow matching pre-training, with the training error depicted in Figure \ref{F_dr_ks}(c), and we obtain $\hat{F}_1 = 0.0099s_{xx}+0.9955u$. Herein, the regularization parameter $\alpha$ is set to 0.01, resulting in the training loss exceeding the validation loss. In stage 2, $\hat{F}_1$ was fine-tuned through the ODESolve prediction, with the corresponding prediction error illustrated in Figure \ref{F_dr_ks}(d), culminating in $\hat{F}_1 = 0.0098s_{xx} + u - 0.0083s + 0.0046$. Herein, we employ the "Euler" method for the initial 120 epochs, incorporating progressively increasing prediction steps. Subsequently, for the remaining 80 epochs, the number of prediction steps is maintained at a constant 20, and we switch to the "Dopri5" method for further training. In stage 3, we maintain $\hat{F}_1$ fixed and employ a GeNN for residual learning, where the optimization objective is $F-\hat{F}_1$ and the prediction error is presented in Figure \ref{F_dr_ks}(e). Herein, the training strategy remains consistent with that of stage 2. In fact, within the temporal-spatial domain of the DR experiment, the influence of $Ks^2$ on the dynamics is minimal. Consequently, this term was not identified in stage 1. However, accurate predictions were achieved through the employment of simple substitute terms in stage 2. Then in the stage 3, the modeling accuracy was further enhanced through residual learning.

\begin{figure} 
	\centering
	\includegraphics[width=0.46\textwidth]{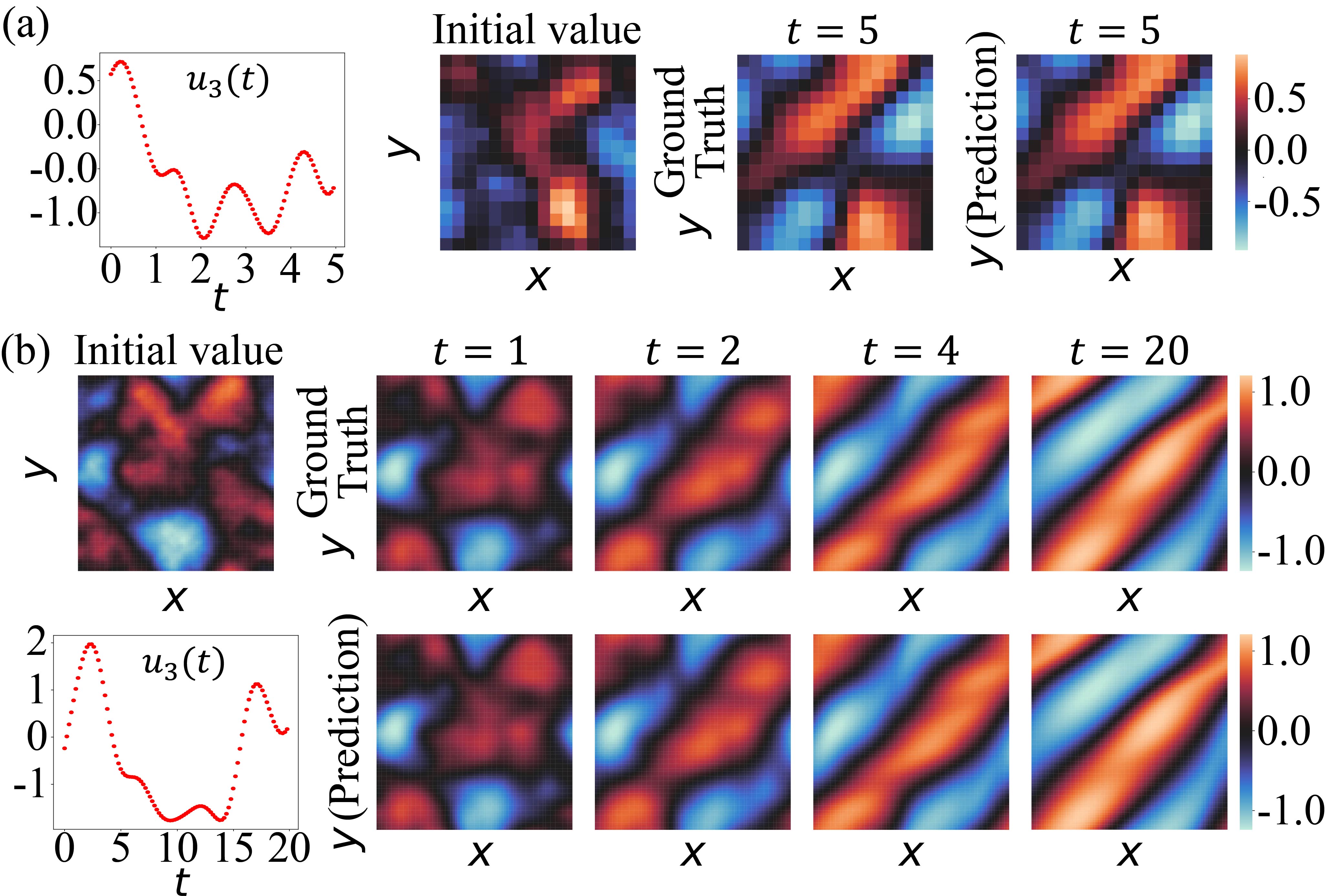}
	\vspace{-0.15cm}
	\caption{Predicting NS systems using SNODEs. (a) The predicted result for training data with spatial resolution $N_x=N_y=16$. (b) The predicted result for testing data with spatial resolution $N_x=N_y=80$.} 
	\label{F_ns}
	\vspace{-0.4cm}
\end{figure}

Finally, we consider a $2$-d NS system for a viscous, incompressible fluid in vorticity form, which reads
\begin{equation}\label{ns_equation}
	\begin{split}
		s_t = \gamma_x s_y - \gamma_y s_x + \nu \Delta s + u(x,y,t),\,\,	\Delta \gamma = -s, \\ 
		\quad (x,y)\in[0,2]^2,\, t\in[0,20],
	\end{split}
\end{equation}
where $\gamma$ represents the stream function, $\Delta$ is the Laplacian operator, and $\nu=0.001$. Additionally, $u(x,y,t) = u_3(t)\times \{0.1\sin[2\pi (x + y)] + \cos[2\pi (x + y)]\}$ is the forcing function, where $u_3(t)$ is a function obtained from a GRF. The system is defined over a square domain with dimensions $[0, 2]^2$, and the time interval is $[0,20]$. 
Under this condition, we can express the stream function as the vorticity, i.e., $\gamma = -\Delta^{-1} s$. In the discrete scenario within the Fourier domain, this corresponds to $\tilde{\gamma} = -1/(\bm{k}^2+\bm{l}^2)\tilde{s}$. To facilitate the training, we augment the first layer of SymNet with $i\bm{k}/(\bm{k}^2+\bm{l}^2)\tilde{s}$ and $i\bm{l}/(\bm{k}^2+\bm{l}^2)\tilde{s}$ in the Fourier domain. Then after training, Figure \ref{F_ns} demonstrates that our framework achieves the accurate operator learning in modeling the underlying dynamics, enabling the precise and stable prediction of system evolutions, even the initial values and parameter functions outside the training set distribution. Additional training details and experimental results for the above parametric PDE systems are provided in Appendix B.4.

\begin{table*}[htbp]
	\centering
	\fontsize{9pt}{8.5pt}\selectfont
	\setlength{\tabcolsep}{4pt}
	\vspace{-0.1cm}
	\begin{tabular}{c|ccc|ccc|ccc}
		\toprule
		Conditions &  \multicolumn{3}{|c}{$N_\text{tr}=100, \, \sigma_\text{n}=0\%$} & \multicolumn{3}{|c}{$N_\text{tr}=1000, \, \sigma_\text{n}=0\%$} & \multicolumn{3}{|c}{$N_\text{tr}=1000, \, \sigma_\text{n}=3\%$}\\
		\midrule
		Experiments &DR &KS &NS &DR &KS &NS &DR &KS &NS\\
		\midrule
		SNODEs &\bf{6e\text{-}4}$\pm$5e\text{-}3 &\bf{0.03}$\pm$0.02 &\bf{0.11}$\pm$0.09 &\bf{3e\text{-}6$\pm$2e\text{-}6}  &\bf{2e\text{-}5$\pm$3e\text{-}4} &\bf{6e-4}$\pm$\bf{5e-3} &\bf{0.53}$\pm$\bf{0.28} &\bf{0.41}$\pm$\bf{0.14} &\bf{1.37}$\pm$\bf{0.67} \\
		DeepONet &27.5$\pm$32.4 &30.2$\pm$24.1 &38.3$\pm$42.4 &0.67$\pm$0.53 &2.56$\pm$1.87 &7.30$\pm$5.24 &3.74$\pm$2.64 &4.39$\pm$4.33 &21.6$\pm$29.8\\
		FNO &30.1$\pm$35.3 &72.1$\pm$42.2 &43.9$\pm$35.1 &1.70$\pm$1.09 &2.90$\pm$3.08 &5.10$\pm$3.54 &2.45$\pm$2.28 &3.93$\pm$2.68 &9.22$\pm$7.63 \\
		PDE-NET &4.21$\pm$3.81 &3.13$\pm$2.66 &8.20$\pm$6.68 &2.04$\pm$1.25 &1.31$\pm$1.42 &3.30$\pm$2.87 &2.28$\pm$1.86 &2.01$\pm$2.42 &3.49$\pm$2.38 \\
		MP-PDE &2.79$\pm$3.24 &4.67$\pm$3.12 &6.65$\pm$3.68 &0.62$\pm$0.28 &1.57$\pm$0.95 &2.26$\pm$1.74 &1.18$\pm$0.61 &1.94$\pm$1.34 &2.85$\pm$2.06 \\					
		E1 (no SymNet) &0.18$\pm$0.12 &0.32$\pm$0.21 &0.35$\pm$0.28 &0.01$\pm$0.03 &0.11$\pm$0.14 &0.08$\pm$0.13 &1.06$\pm$0.84 &1.39$\pm$0.66 &1.55$\pm$1.27\\			
		E2 (no GeNN) &0.41$\pm$0.62 &0.08$\pm$0.04 &0.68$\pm$0.27 &0.17$\pm$0.15 &5e-5$\pm$6e\text{-}4 &3e-3$\pm$8e\text{-}3 &0.93$\pm$1.09 &0.82$\pm$0.94 &1.67$\pm$1.52\\	
		\bottomrule
	\end{tabular}%
	\vspace{-0.1cm}
	\caption{The prediction MSE ($\pm$ two standard deviations) under different training set sizes $N_\text{tr}$ and noise levels $\sigma_\text{n}$. Here, we consider zero-mean Gaussian noise with a standard deviation of $\sigma_\text{n}$ times the mean absolute value of the training data.}
	\label{T_compare}%
	\vspace{-0.2cm}
\end{table*}% 

\subsection{Comparative Analysis and Ablation Studies}
For the fair comparison, we conduct a comparative analysis with several standard baselines, namely DeepONet \cite{Lu2019DeepONetLN}, FNO \cite{Li2020FourierNO}, PDE-NET \cite{long2019pde}, and MP-PDE \cite{Brandstetter2022MessagePN}. Additionally, we conducted ablation experiments on SNODEs by separately removing SymNet and GeNN, denoted as E1 (no SymNet) and E2 (no GeNN) respectively. Then we employ the high-precision pseudospectral method to solve parametric PDEs and obtain the dataset. In particular, to increase the task difficulty, we set the output scale of the sampling function to be larger than that of the training set and randomly generated initial values (see Appendix B.2 for details).

The experimental results, which are presented in Table \ref{T_compare}, demonstrate that our framework is capable of accurately modeling system dynamics even in scenarios where training data is limited and noisy. Consequently, the SNODEs outperform the DeepONet and FNO methods in terms of interpretability and extrapolation capability, particularly in the NS experiment with random initial values. This is because SNODEs sufficiently leverage additional physical priors by adding the partial derivative terms into $\bm{F}$. Compared to the PDE-NET and MP-PDE methods, our framework enables efficient symbolic regression over a larger search space that includes parametric functions. This is due to the joint use of three-stage learning and ODESolve strageties, enabling the more precise modeling of spatiotemporal dynamics from limited noisy data. Furthermore, results from ablation experiments E1 and E2 clearly demonstrate that these two components mutually enhance each other, with SymNet enhancing the model's interpretability and generalization abilities, while GeNN strives for greater modeling precision.

%Additionally, our framework also achieves more stable and long-term prediction performance by selecting superior ODE solvers, such as \texttt{dopri5}.

\begin{table}[htbp]
	\vspace{-0.1cm}
	\centering
	\fontsize{9pt}{8pt}\selectfont
	\setlength{\tabcolsep}{1.9pt}
	\vspace{-0.05cm}
	\begin{tabular}{c|cccccc}
		\toprule
		Config &SNODEs &NODE &DeepONet &FNO &PDE-NET &MP-PDE \\
		\midrule
		DR &221 &1580 &372 &557 &430 &943 \\
		KD &489 &2920 &407 &649 &513 &1017 \\
		NS &824 &12476 &694 &1120 &902 &2389 \\
		\bottomrule
	\end{tabular}%
	\vspace{-0.15cm}
	\caption{ Comparison of training times using different methods. Here, the unit of measurement is seconds.}
	\label{T_computation_cost}%
	\vspace{-0.2cm}
\end{table}%

As an advancement of the NODEs, our SNODEs framework employs the flow matching pre-training strategy, effectively optimizing the high computational cost associated with classical NODE methods. To assess the efficiency of our proposed method, we conduct the runtime comparisons with the baselines in the DR, KS, and NS experiments. The results are displayed in the Table \ref{T_computation_cost} of the attached file. It demonstrates that our framework not only achieves the higher precision but also maintains the good training efficiency. However, it is worth noting that the computation cost during the inference process is contingent upon the choice of the ODE solver, with higher precision numerical solvers potentially requiring longer computation time.

\section{Concluding Remarks}
In this work, we present the SNODEs framework for learning the operator mapping from the parameter functions to system states in parametric dynamical systems. Specifically, this framework establishes the SCDNNs which effectively integrate the strengths of symbolism and connectionism. It initially employs SymNet to learn potential underlying equations, followed by utilizing GeNN for residual learning, thereby achieving a more accurate model of the physical world. To enhance the robustness of our method, we also propose the three-stage and adaptive training strategy that simultaneously improves the stability and efficiency of modeling unknown dynamics. 

Moreover, by leveraging the Fourier transform to incorporate potential partial derivative terms into the first layer of SymNet, SNODEs enable accurate and resolution-invariant modeling of the PDE systems at any given time and on any spatial grid. We evaluate SNODEs method across multiple representative ODE and PDE systems. Experimental results demonstrate that, compared to the standard baselines, our framework not only achieves higher prediction accuracy but also exhibits superior extrapolation capabilities, interpretability, and robustness. Consequently, our framework can be applied to a broader range of scientific problems and real-world systems.

However, our approach has certain limitations that warrant further investigation in the future. For example, the Fourier transform within the SNODEs framework is primarily suited for uniformly spaced spatial data. Therefore, for PDE systems encompassing complex geometric domains, integration with advanced strategies (such as non-uniform Fourier transform or other numerical methods) is necessary. In addition, when the underlying dynamics become increasingly complex, deep SymNet may exhibit instability during the training process. Therefore, incorporating more sophisticated symbolic regression approaches, such as KAN \cite{liu2024kan}, into the SNODEs framework represents a potential and significant direction for future exploration.

\bibliography{aaai25}

\end{document}